# Integrating a Digital Twin Concept in the Zero Emission Sea Transporter (ZEST) Project for Sustainable Maritime Transport using Stonefish Simulator

Michele Grimaldi * [†], Carlo Cernicchiaro[‡] [‖], George Rossides[‡], Angelos Ktoris[§], Elias Yfantis [§] [**],
and Ioannis Kyriakides[¶] [**]

*Computer Vision and Robotics Institute
University of Girona, Girona, Spain

[†]School of Engineering & Physical Sciences
Heriot-Watt University
Email: m.grimaldi@hw.ac.uk

[‡]Marine Robotics Innovation Centre
Cyprus Marine and Maritime Institute, Larnaca, Cyprus

[§]Marine & Offshore Science, Technology & Engineering Centre
Cyprus Marine and Maritime Institute, Larnaca, Cyprus

[¶]Marine Technology Division
Cyprus Marine and Maritime Institute, Larnaca, Cyprus
Email: carlo.cernicchiaro, george.rossides,
angelos.ktoris, elias.yfantis, ioannis.kyriakides @cmmi.blue

[‖]Computational Cognition Lab
Open University of Cyprus, Nicosia, Cyprus

[**] Department of Engineering, University of Nicosia, Nicosia, Cyprus
Email: yfantis.e, kyriakides.i @unic.ac.cy

## I. INTRODUCTION

In response to stringent emission reduction targets imposed by the International Maritime Organization (IMO) and the European Green Deal's Fit for 55 legislation package, the maritime industry has shifted its focus towards decarbonization. While significant attention has been placed on vessels exceeding 5,000 gross tons (GT), emissions from coastal and short sea shipping, amounting to approximately 13% of global shipping transportation and 15% within the European Union (EU), have not received adequate consideration. This abstract introduces the Zero Emission Sea Transporter (ZEST) project, designed to address this issue by developing a zero-emissions multi-purpose catamaran for short sea routes, shown in Figure 1.

The ZEST [1] is envisioned as a vessel and a multifaceted research platform with a broad spectrum of applications. Its objectives encompass supporting the research activities of the CMMI Cyprus Marine and Maritime Institute and its vast partners network, serving as a testing ground for industrial technologies, and aiding CMMI's vocational education and training initiatives. This multi-functional project is structured into distinct activities, each addressing critical aspects of sustainable maritime transport and education and training

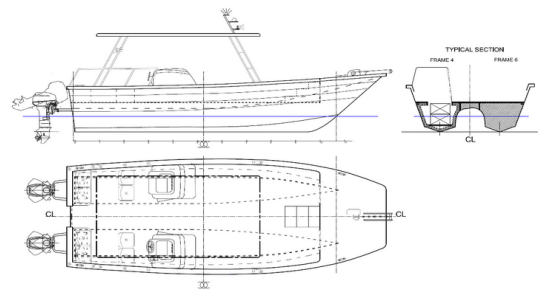

Fig. 1. General Arrangement Plan of ZEST

activities:

- **Decarbonization Technologies**: ZEST provides a test bed for various decarbonization technologies, methodologies, and best practices. It is a platform for evaluating alternative propulsion systems, including fuel cells and hybrid systems and testing various alternative fuels in conventional internal combustion engines, such as gaseous and liquid bio-fuels and blends with fossil fuels.
- **Navigational Autonomy**: The project involves designing, testing, and validating algorithms for navigational autonomy. These algorithms encompass path planning, collision avoidance, and real-time decision-making, often incorporating advanced computer vision techniques for environmental recognition and depth perception.
- **Deployment Platform for Mechatronic Systems**: ZEST is a versatile deployment platform for autonomous or semi-autonomous vehicles, including underwater, surface, and aerial platforms. This capability enables various missions, such as seabed mapping, surveillance, inspections, and collaborative swarm tasks.
- **Environmental Monitoring**: The vessel is equipped for environmental monitoring, conducting baseline surveys, and long-term tracking through audiovisual techniques and sensors. It facilitates the collection of data on biological parameters (e.g., mapping priority habitats and recording endangered species) and physicochemical parameters (e.g., temperature, salinity, chlorophyll). Additionally, it plays a pivotal role in pollution detection from both inorganic and organic sources.
- **Educational and Training Platforms**: ZEST serves as an educational and training platform, offering Vocational Education and Training (VET) courses in diverse fields, including naval architecture and marine engineering, communication engineering, artificial intelligence and signal processing, and robotics engineering. It also supports undergraduate and postgraduate studies through collaboration with local and international universities.

To optimize the vessel's energy management, a central Power and Energy Management and Monitoring System (PEMMS) has been developed. PEMMS has intelligent algorithms, sensors, actuators, and user interfaces to optimize power consumption. These algorithms monitor the vessel's status and identify malfunctions or unusual behavior, ensuring efficient energy usage. Furthermore, ZEST has a fully electric propulsion system, ensuring a carbon-free and zero-emission operation. At the same time, incorporating solar photovoltaic (PV) panels contributes to the goal of harnessing renewable energy sources. The vessel's design minimises hull resistance and drag, ensuring efficient energy utilization and extended range. Its catamaran configuration offers maximal internal volume for the installation of batteries, and the ample deck space can accommodate solar panels, partially covering energy needs. The vessel's design supports low cruising speeds ranging from 5 to 11 knots, with an 80 kWh battery set to achieve the required speed and range.

## II. Related Work

### A. Guidance System

Guidance systems allow the steering of autonomous ships along predefined paths, ensuring optimal navigation. Different techniques were explored in the literature. Geometric techniques such as Line of Sight (LOS) guidance laws are widely employed. [8] utilized nonlinear controllers, demonstrating stability through backstepping control. [16] introduced dynamic look-ahead LOS, enhancing convergence. [4] extended LOS with integral action for drift compensation, validated by [21]. Pure Pursuit (PP) guidance, analyzed by [18], offers an alternative. [24] optimized path following by combining PP and LOS. Constant Bearing (CB) guidance, presented by [7], finds applications in both collision avoidance and targeted navigation. [19] utilized adaptive neural networks for CB guidance. Vector Field (VF) guidance, initially for air vehicles [17], was adapted in [23] for marine surface ships. Time-varying VF guidance, proposed by [22], adds stability. In our simulations we used a PID controller integrated with Virtual Target Guidance (VTG), as proposed by [27], which simplifies the problem by introducing a moving virtual target satisfying impact time and angle constraints and represents a practical and effective solution for navigating autonomous ships in dynamic environments.

### B. Path Planning

Artificial Potential Field (APF) is a well-explored global and local path planning technique [12]. This method has been optimized and modified several times. In [13], the environment is represented as a two-dimensional coordinate system, with forces calculated based on the Euclidean distances between the robot, target, and obstacles. The methodology enhances traditional APF by introducing a distance term to the repulsive field function, addressing local minima and path oscillations. This integration ensures that the target point is the global minimum of the potential field, improving path smoothness and reachability. Decision tree algorithms adapt the APF to different scenarios, focusing on parameters like repulsive gain coefficient, offset angle of the resultant force, and velocity. These modifications enable real-time, accurate decision-making in path planning. Simulation results demonstrate a 50% reduction in planning time and a 43.3% increase in path smoothness compared to the traditional APF method.

APFs have also been used on USVs. In [20], the path planning for USVs is addressed using a Predictive APF (PAPF) method. The predictive component allows the USV to anticipate future environmental conditions and adjust its path accordingly. While repulsive potential is static, depending solely on the obstacle's position and remaining constant regardless of the USV's location, predictive potential dynamically varies based on the relative positions of the USV and obstacles. Predictive potential aims to foresee feasible paths, adapting to blockages by generating lateral forces for deflection, thus avoiding local minima and ensuring smoother navigation. Furthermore, the PAPF is extended to address moving obstacle avoidance through the Moving Obstacle Prediction algorithm, which anticipates future obstructions and adjusts the USV's trajectory accordingly.

The APF method can equate the movement of a ship in an actual environment to that in an artificial attractive and repulsive potential energy field, as in [11], where environments are treated as potential energy fields where other ships and obstacles exert repulsive potential energy, generating a force that keeps the ship away from them. Conversely, the target point exerts an attractive potential energy, guiding the ship towards it. The attractive potential field is determined by the relative position of the ship and the target, with the attractive force inversely proportional to the square of their distance. Repulsive potential energy fields are created around other ships and obstacles, generating forces that push the ship away within a certain influence radius. The combination of these forces, governed by the APF method, aims to guide the ship safely through its environment, complying with COLREGs, especially when one ship fails to take collision avoidance measures.

A modified APF algorithm for autonomous trajectory planning in ships or USVs is used in [14]. The algorithm is tailored for dynamic environments, addressing stationary and moving hazards while complying with the COLREGs. The attractive potential field in the model is a combination of relative position and velocity fields, adapting to dynamic goals. The repulsive potential field function is subdivided and considers factors like stationary and dynamic obstacles, multi-ship collision avoidance, and close-range emergency actions. This enhanced APF model ensures safe passage around obstacles and aligns with COLREGs by directing avoidance actions to the starboard for dynamic ships.

In [15], firstly, the collision risk is determined by two factors: Checking Criterion of Collision Risk (CC) and Checking Range of Collision Risk (CR). The methodology employs pos and vos for the own ship's (OS) position and velocity and pts and its target ships (TS). The algorithm expands the TS's safe boundary based on the OS and TS domain radius and an allowable safe distance, resulting in a dynamic collision risk circle. This circle's radius accounts for visibility conditions and operator settings. Then, a modified APF integrates decision-making, considering both stationary and moving TSs. It introduces a subdivided repulsive potential field function for collision avoidance (CA) actions, considering the relative positions and velocities of the OS and TSs. The total repulsive force drives the OS to alter course as per COLREGs, ensuring safety in various navigational scenarios.

For obstacle avoidance, ZEST employs an approach inspired by the PAPF in [20], combining it with a behavior tree, similarly to [13].

### C. Simulation Environments in Collision Avoidance Studies

The literature surrounding ship simulation is rich with endeavours to model vessel dynamics and enhance collision avoidance systems. In [26] the authors use a MATLAB-Simulink environment to simulate a 3DOF ship and their Optimal Path Planning (OPP) in the presence of fixed obstacles and other ships. In [25] the authors use MATLAB and Unity3D to simulate jointly, establish 3D navigation environment in Unity3D, and complete the training of their CICA (COLREGs intelligent collision avoidance) algorithm which combines the feature extraction ability of deep convolution neural network with the decision-making ability of reinforcement learning. In [2] the authors developed their own simulator VRX (Virtual RobotX) [3] based on Gazebo. In their simulation each vessel has its own global and local motion planners complying with COLREGS while a Scenario-based Model Predictive Control (SBMPC) approach [10] is used for collision avoidance and a dynamic Bayesian network (DBN) is utilized to estimate other ships' intentions.

## III. METHODOLOGY

### A. Integration with Stonefish Simulator and Digital Twin Concept

Integrating the ZEST vessel with the Stonefish simulator [5] holds paramount significance, as it showcases the potential of the digital twin concept and represents a pioneering feat. For the first time, a vessel of these dimensions, a substantial 10 meters long, has been successfully simulated within the Stonefish environment. Stonefish, a powerful C++ library, enables realistic simulations of marine robots and offers compatibility with the Robot Operating System (ROS) through the stonefish-ros package. Stonefish ensures precise real-time sensor positioning, allowing accurate mapping and environmental monitoring while aligning the simulator with real-world conditions. It validates the vessel's performance within a controlled environment, using versatile sensors like multi-beam echo sounders, profilers, cameras, and depth cameras to collect comprehensive data and gain a deep understanding of the vessel's behavior. In the context of the ZEST vessel's integration with the Stonefish simulator, achieving a stable simulation scenario is contingent upon the meticulous preparation of geometry files. The critical nature of this requirement stems from the fact that the geometry of models directly influences the computation of physics within the simulator. When using mesh data to create simulated bodies, Stonefish allows for separate visual and physical geometry definitions. Notably, only the physical geometry significantly affects the simulation and requires thorough preprocessing. This segregation between simulation and visualization provides users with the flexibility to input meshes of varying resolutions and levels of detail. Consequently, it optimises simulation accuracy while concurrently ensuring high-quality rendering. The author uses Blender [6] to prepare the geometry for the simulations. In the simulation, the vessel is equipped with various sensors, such as:

- **Inertial Measurement Unit (IMU)** to get orientation, angular velocities, and linear accelerations;
- **Global Positioning System (GPS)** to get the position of the vehicle
- **Compass** to get the heading of the vehicle

### B. VTG Trajectory Tracking

The ZEST vessel operates as a differential drive, where its two motors produce identical thrust for linear movements and varied thrust for turning maneuvers. VTG is critical in this setup, as it helps calculate the optimal linear and angular velocities needed for the vessel to reach its next target position effectively. The PID translate these into differential thrust commands, generating control signals to adjust the thrust of the two motors. This approach ensures that ZEST can efficiently navigate straight paths and complex turns, fluidly adapting its motion to the moving virtual target. Implementing VTG in the ZEST project offers several advantages since it enhances the vessel's path-tracking accuracy, ensuring that ZEST follows the planned route and, due to its adaptability, it is particularly beneficial in changing sea conditions, allowing ZEST to adjust its course dynamically in response to environmental factors or obstacles.

*C. COLREGs with Behavior Trees and Artificial Potential Field*

This section discusses the collision avoidance strategy employed by ZEST based on the COLREGs.

In a simulated scenario, the position, orientation, and linear and angular velocity of the other boat are known, providing the necessary information to determine which COLREGs rule should be applied. In the case of ZEST, the behavior tree serves as a hierarchical decision-making structure that governs the system's actions. The utilization of behavior trees in the context of maritime traffic simulations, as highlighted by [9], provides a modular and flexible approach to defining and controlling the behavior of autonomous agents.

Our hierarchical control structure is shown in Figure 2. The following elements are used:

- Root (green rectangle): Represents the entry point of the BT. It has a control node as a single child.
- Sequence Node ($\rightarrow$ in yellow rectangle): A control node that executes its children in sequence. It fails if any child fails; otherwise, it succeeds. If one of its children returns running, the Sequence node keeps ticking it (returning running as well) until the child does not succeed or fail. If it is recursive (R $\rightarrow$), the node will restart the ticking of all children if one returns running.
- Fallback Node (? in yellow rectangle): A control node that executes its children in order until one succeeds. Succeeds if any child succeeds; otherwise fails. If a child of the Fallback node returns a running state, the node continues to tick that child (also returning a running state) until the child succeeds or fails. In the case of recursive Fallback (R ?), the node will initiate a restart if one of its children returns running.
- Condition Node (red eclipse): Represents a condition or check that returns either true or false.
- Leaf Node (red rectangle with circular angles): Represents a terminal action or condition in the behavior tree. It can return success (the action succeeded), failure (the action failed), or running (the action is still in progress).

The usage of BT guarantees the explainability of the ZEST behavior.

The first step involves checking if the trajectory is clear from any current or dynamic obstacles that could intersect its path. If the trajectory is obstacle-free, ZEST proceeds directly toward its goal. This decision is made by ticking the "MoveToTarget" leaf node in the BT. This leaf node returns "RUNNING" until ZEST reaches its target, returning "SUCCESS".

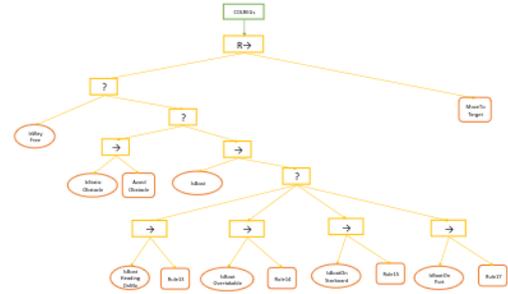

Fig. 2. BT regulating COLREGs to be applied

If the "MoveToTarget" leaf node returns "RUNNING", indicating the presence of obstacles, the BT goes back and evaluates the surroundings using the Condition "IsWayFree". If ZEST's trajectory is still clear from obstacles, the "MoveToTarget" leaf node is ticked again. However, if obstacles are detected, the system tries to identify whether the obstacle is a static obstacle or another boat.

If the obstacle is identified as another marine vehicle, the corresponding COLREGs rule is activated by ticking the respective leaf node. This leaf node returns "RUNNING" until the rule is wholly followed.

According to the active boat rule, ZEST dynamically builds a potential field around the other boat. This potential field shapes ZEST's behavior, ensuring it maintains a minimum safety distance equal to twice the beam (width) of ZEST. The approach considers various scenarios based on the relative orientation and motion of the other boat.

- If the other boat is overtakable (Rule 13) or approaching ZEST head-on (Rule 14), ZEST slows down and aims to reroute to a more convenient path, either to the port or starboard side based on the position of the other boat. This manoeuvre helps avoid potential collisions and ensures safety while passing the obstacle.
- If the other boat is on the starboard side (Rule 15), the PAPF directs ZEST's trajectory to reroute on the starboard side, effectively passing behind the beam of the other boat. This maneuver aligns with the established COLREGs rule for right-of-way.
- If the other boat is on the port side (Rule 17), ZEST maintains its current route, considering that the COLREGs do not require significant changes in this scenario.

IV. RESULTS

*A. VTG Trajectory Tracking Performance*

In evaluating VTG Trajectory Tracking for the ZEST project, a distinctive 8-shaped trajectory was chosen as the test path. This trajectory presents a comprehensive challenge for the navigation system, combining straight-line navigation with complex turning maneuvers. The objective was to assess the effectiveness of our VTG algorithm in guiding the ZEST vessel along this intricate path. The vessel was tasked to autonomously navigate the trajectory, controlled entirely by the VTG algorithm through the PID system. As shown in Figure 3, the results demonstrated the capability of our VTG control algorithm to manage complex navigational tasks. The ZEST vessel followed the 8-shaped trajectory throughout the test, closely adhering to the predefined path. Notably, during the execution of the trajectory's turns — the more challenging aspect of the path — the vessel's differential drive system effectively adjusted the thrust of the two motors, allowing for smooth and precise navigation around the curves.

### B. Collsion avoidance COLREGs compliant

The obstacle avoidance behavior of ZEST, compliant with COLREGs, was evaluated using simulation in Stonefish. Two models of ZEST were used in the experiments: a white ZEST and a red ZEST, with the trajectories described in Figure 4, respectively, in blue and red. It is important to note that only the white ZEST was commanded during the experiments, while the red ZEST blindly moved forward. To test the COLREGs compliance with the white ZEST, various scenarios were considered by placing the red ZEST in different positions and orientations. The results of the evaluation are as follows:

- Overtaking the red ZEST: In this scenario, the white ZEST overtook the red ZEST. The white ZEST successfully applied COLREG Rule 13 (Figure 5), which requires the overtaking vessel (white ZEST) to maintain its course and speed while ensuring the safe passing of the overtaken vessel (red ZEST). The white ZEST adjusted its speed and position to execute the overtaking manoeuvre.
- Heading on the white ZEST, right in front of it: In this case, the red ZEST approached the white ZEST head-on, coming directly towards it. The white ZEST effectively applied COLREG Rule 14 (Figure 6), which mandates the giving-way vessel (red ZEST) to alter course or speed to avoid collision. The white ZEST slowed down and rerouted, ensuring the safe passage of both ships.
- Crossing from the starboard of the white ZEST at 90 degrees (Figure 7): In these scenarios, the red ZEST approached the white ZEST from the starboard side. The white ZEST successfully applied COLREG Rule 15, correctly rerouting to the starboard side and passing behind the beam of the red ZEST to avoid a potential collision.
- On the port side of the white ZEST: In this scenario, the red ZEST was placed on the port side of the white ZEST. The white ZEST effectively applied COLREGs Rule 17 (Figure 8), which stipulates that the stand-on vessel (white ZEST) should maintain its course and speed when a ship (red ZEST) is overtaking it on the port side. The white ZEST continued its course without significant changes, respecting the regulations.

## V. CONCLUSION

This paper presented a comprehensive study on integrating a Digital Twin concept in the ZEST project, emphasizing sustainable maritime transport using the Stonefish Simulator. The successful implementation of the VTG algorithm and adherence to COLREGs through behavior trees and APF has been key in realizing ZEST's autonomous navigation capabilities.

The ZEST project integrates advanced navigation and simulation technologies in maritime transport. The project aligns with global efforts towards decarbonization and sets a benchmark for future developments in autonomous maritime systems. The results from this study offer valuable insights into the potential of digital twin technologies in enhancing maritime operations' safety, efficiency, and sustainability. Future work could explore the scalability of these technologies in larger vessels and more complex maritime scenarios.

## VI. ACKNOWLEDGMENTS

This research was funded by CMMI Cyprus Marine and Maritime Institute. CMMI was established by the CMMI/MaRITeC-X project as a "Center of Excellence in Marine and Maritime Research, Innovation and Technology Development" and has received funding from the European Union's Horizon 2020 research and innovation program under grant agreement No. 857586 and matching funding from the Government of the Republic of Cyprus.

## REFERENCES

[1] G. Mallouppas C. Ioannou S. Damianou F. Gavriel T. Kounoudes K. Michail A. Ktoris, E. Yfantis. The zest project: Design of a smart, zero-emission, multi-purpose vessel. 2023.
[2] Mustafa Bayrak and Haluk Bayram. Colreg-compliant simulation environment for verifying usv motion planning algorithms. In *OCEANS 2023 - Limerick*, pages 1–10, 2023.
[3] Brian Bingham, Carlos Agüero, Michael McCarrin, Joseph Klamo, Joshua Malia, Kevin Allen, Tyler Lum, Marshall Rawson, and Rumman Waqar. Toward maritime robotic simulation in gazebo. In *OCEANS 2019 MTS/IEEE SEATTLE*, pages 1–10. IEEE, 2019.
[4] Even Borhaug, A. Pavlov, and Kristin Y. Pettersen. Integral los control for path following of underactuated marine surface vessels in the presence of constant ocean currents. In *2008 47th IEEE Conference on Decision and Control*, pages 4984–4991, 2008.
[5] Patryk Cieślak. Stonefish: An advanced open-source simulation tool designed for marine robotics, with a ros interface. In *OCEANS 2019 - Marseille*, jun 2019.
[6] Blender Online Community. *Blender - a 3D modelling and rendering package*. Blender Foundation, Stichting Blender Foundation, Amsterdam, 2018.
[7] Thor Fossen. *Handbook of Marine Craft Hydrodynamics and Motion Control*. 04 2021.
[8] Thor Fossen, Morten Breivik, and Roger Skjetne. Line-of-sight path following of underactuated marine craft. 09 2003.
[9] Volker Gollücke, Daniel Lange, Axel Hahn, and Sören Schweigert. Behavior tree based knowledge reasoning for intelligent vessels in maritime traffic simulations. In *ECMS*, pages 105–113, 2018.
[10] IB Hagen, DKM Kufoalor, TA Johansen, and EF Brekke. Scenario-based model predictive control with several steps for colregs compliant ship collision avoidance. *IFAC-PapersOnLine*, 55(31):307–312, 2022.
[11] Dengjun Huang, Zhengjiang Liu, and Xin Wang. Artificial potential field and ship maneuverability based collision avoidance path planning. In *2020 Chinese Automation Congress (CAC)*, pages 2490–2495. IEEE, 2020.

[12] Oussama Khatib. Real-time obstacle avoidance for manipulators and mobile robots. *The international journal of robotics research*, 5(1):90–98, 1986.

[13] Xin Lin, Zhan-Qing Wang, and Xu-Yang Chen. Path planning with improved artificial potential field method based on decision tree. In *2020 27th Saint Petersburg International Conference on Integrated Navigation Systems (ICINS)*, pages 1–5. IEEE, 2020.

[14] Hongguang Lyu and Yong Yin. Ship's trajectory planning for collision avoidance at sea based on modified artificial potential field. In *2017 2nd International conference on robotics and automation engineering (ICRAE)*, pages 351–357. IEEE, 2017.

[15] Hongguang Lyu and Yong Yin. Colregs-constrained real-time path planning for autonomous ships using modified artificial potential fields. *The Journal of navigation*, 72(3):588–608, 2019.

[16] Lúcia Moreira, Thor I. Fossen, and C. Guedes Soares. Path following control system for a tanker ship model. *Ocean Engineering*, 34(14):2074–2085, 2007.

[17] D.R. Nelson, D.B. Barber, T.W. McLain, and R.W. Beard. Vector field path following for small unmanned air vehicles. In *2006 American Control Conference*, pages 7 pp.–, 2006.

[18] Fotis A. Papoulias. On the Nonlinear Dynamics of Pursuit Guidance for Marine Vehicles. *Journal of Ship Research*, 37(04):342–353, 12 1993.

[19] Raja Rout, Rongxin Cui, and Zhengqing Han. Modified line-of-sight guidance law with adaptive neural network control of underactuated marine vehicles with state and input constraints. *IEEE Transactions on Control Systems Technology*, 28(5):1902–1914, 2020.

[20] Jia Song, Ce Hao, and Jiangcheng Su. Path planning for unmanned surface vehicle based on predictive artificial potential field. *International Journal of Advanced Robotic Systems*, 17(2):1729881420918461, 2020.

[21] Lili Wan, Yixin Su, Huajun Zhang, Binghua Shi, and Mahmoud S. AbouOmar. An improved integral light-of-sight guidance law for path following of unmanned surface vehicles. *Ocean Engineering*, 205:107302, 2020.

[22] Haitong Xu, Thor Fossen, and Carlos Guedes Soares. Uniformly semiglobally exponential stability of vector field guidance law and autopilot for path-following. *European Journal of Control*, 53:88–97, 05 2020.

[23] Haitong Xu and C. Guedes Soares. Vector field path following for surface marine vessel and parameter identification based on ls-svm. *Ocean Engineering*, 113:151–161, 2016.

[24] Haitong Xu and Carlos Guedes Soares. An optimized energy-efficient path following algorithm for underactuated marine surface ship model. *The International Journal of Maritime Engineering*, 160:A413 – A423, 10 2018.

[25] Xinli Xu, Yu Lu, Xiaocheng Liu, and Weidong Zhang. Intelligent collision avoidance algorithms for usvs via deep reinforcement learning under colregs. *Ocean Engineering*, 217:107704, 12 2020.

[26] Raphael Zaccone. Colreg-compliant optimal path planning for real-time guidance and control of autonomous ships. *Journal of Marine Science and Engineering*, 9(4), 2021.

[27] Zhihong Zhang, Kemao Ma, Gongping Zhang, and Liang Yan. Virtual target approach-based optimal guidance law with both impact time and terminal angle constraints. *Nonlinear Dynamics*, 107, 03 2022.

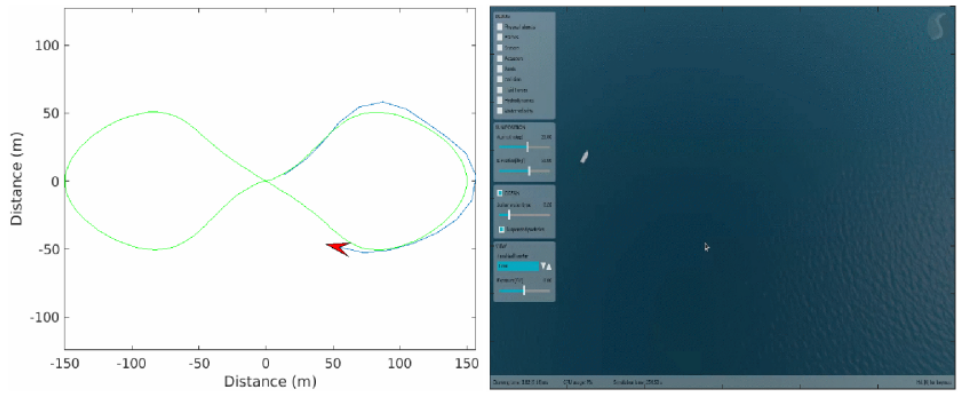

Fig. 3. Path tracking simulations in StoneFish simulator (right). The real-time control of the vessel and the visualization of the vessel trajectory and the reference path were performed using MATLAB (left).

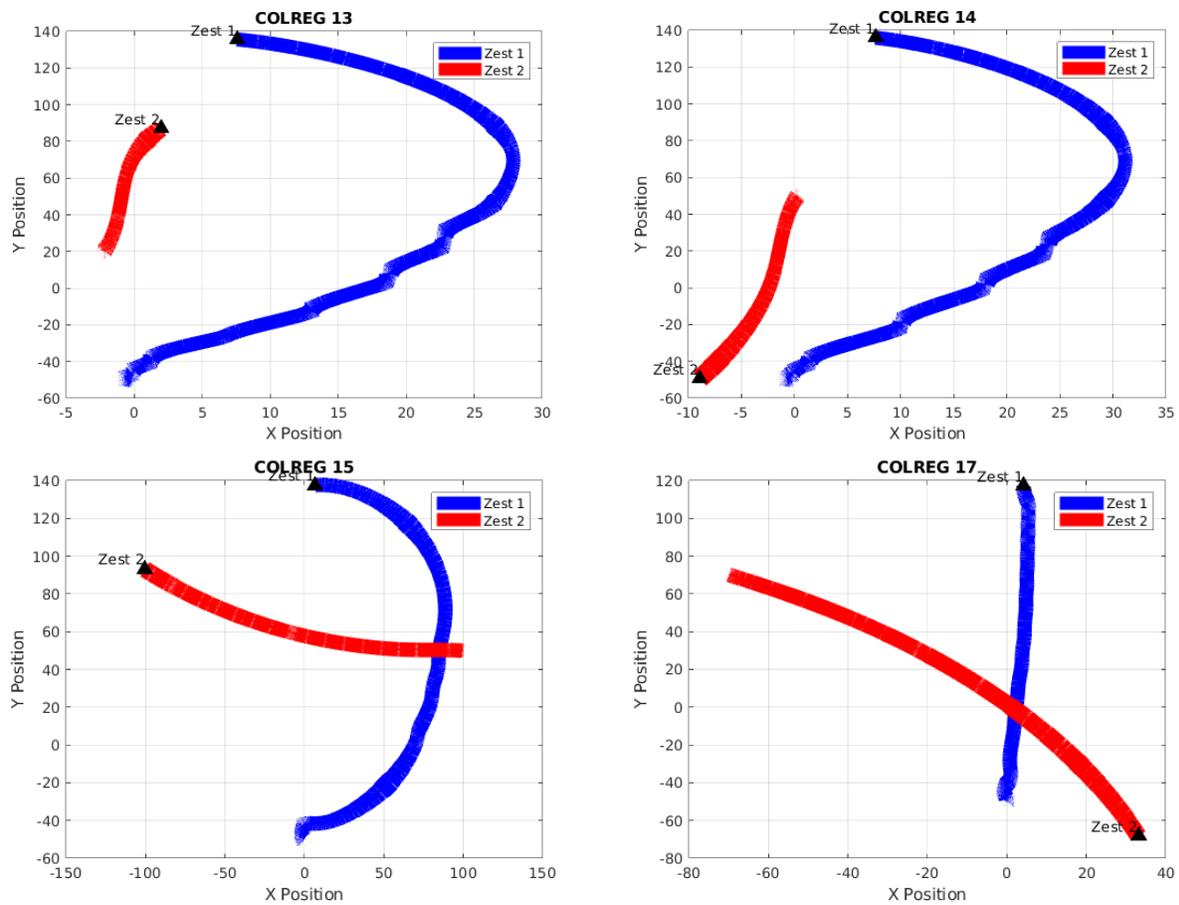

Fig. 4. Graphic of COLREGs 13, 14, 15 and 17. In red, the trajectory of the ZEST moving blindly forward; in blue, the controlled ZEST applying COLREGs

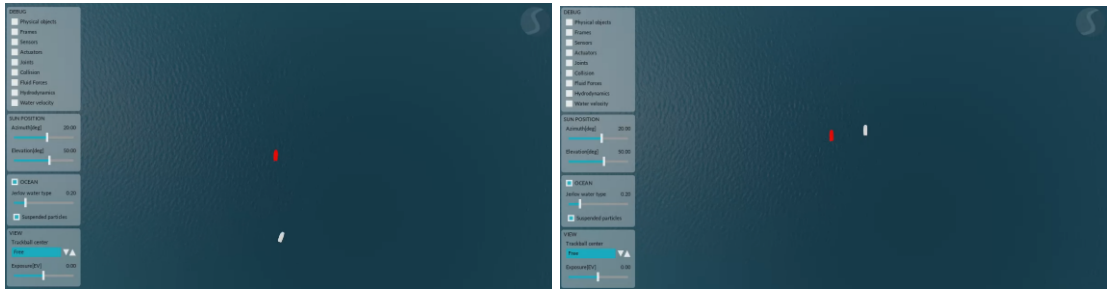

Fig. 5. Rule 13 - Overtaking: this rule is in effect when a vessel approaches another from more than 22.5 degrees above its beam. The responsibility to avoid collision lies with the overtaking vessel

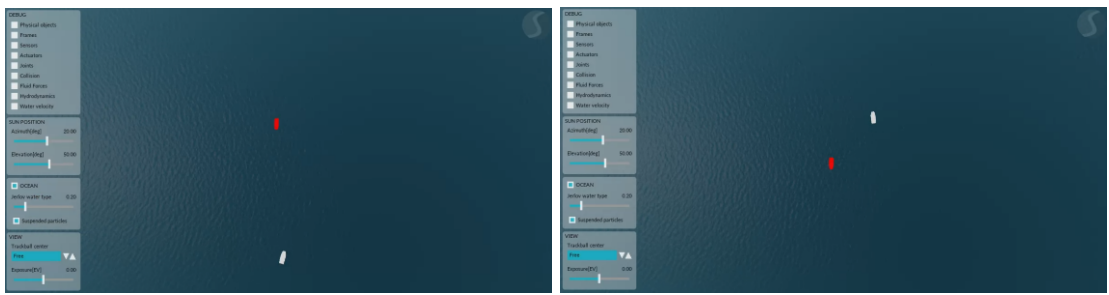

Fig. 6. Rule 14 - Head-On Situation: both vessels should alter their course to starboard to pass on the port side of each other, emphasizing mutual responsibility in head-on situations

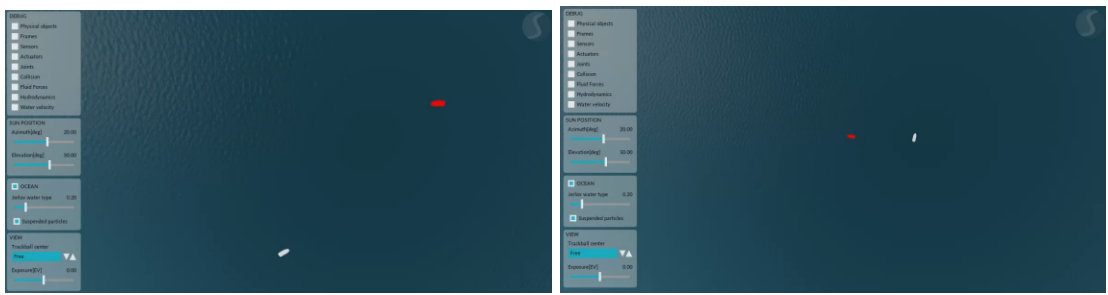

Fig. 7. Rule 15 - Crossing Situation: the vessel with the other on its starboard side must give way and avoid crossing ahead. The other, as the stand-on vessel, should maintain its course and speed

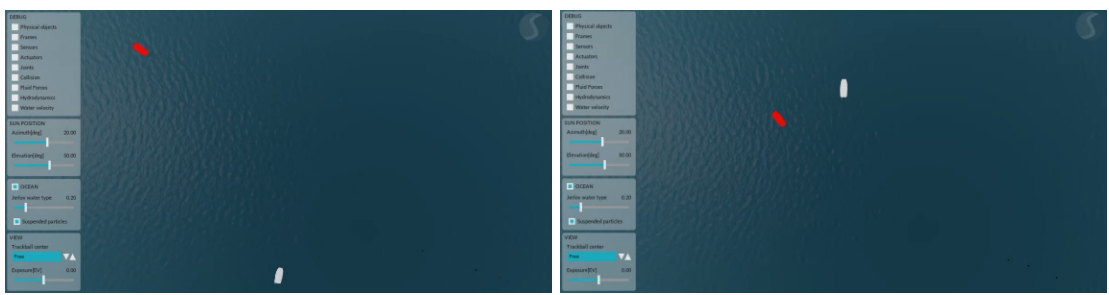

Fig. 8. Rule 17 - Action by Stand-On Vessel: the stand-on vessel should maintain its course and speed but take action to avoid collision if necessary, especially if the give-way vessel fails to act appropriately